\documentclass[journal]{IEEEtran}
\usepackage{xcolor}
\usepackage{algorithmicx}
\usepackage{algpseudocode}
\usepackage{hyperref}
\usepackage{amssymb}
\usepackage{multicol}
\usepackage{subcaption}
\usepackage{graphicx}
\usepackage{lipsum}
\usepackage{soul}
\usepackage{cite}
\usepackage{tabularx}
\usepackage{array}
\usepackage{multirow}
\usepackage{fancyhdr}
\usepackage{graphicx}
\usepackage{amsmath}

\newcolumntype{L}{>{\raggedright\arraybackslash}X}
\algrenewcommand\algorithmicindent{0.8em}%

\algtext*{EndIf}

\DeclareMathOperator*{\argmin}{argmin}
\DeclareMathOperator*{\argmax}{argmax}

\begin{document}

\title{Multi-residual Mixture of Experts Learning for Cooperative Control in Multi-vehicle Systems}

\author{Vindula Jayawardana, Sirui Li, Yashar Farid, Cathy Wu
\thanks{Vindula Jayawardana is with the Massachusetts Institute of Technology, Cambridge, MA 02139, USA.
\tt\small{vindula@mit.edu}}
\thanks{Sirui Li is with the Massachusetts Institute of Technology, Cambridge, MA 02139, USA.
\tt\small{siruili@mit.edu}}
\thanks{Yashar Farid was with Toyota InfoTech Labs when this work was conducted.
\tt\small{y.zeinali@gmail.com}}
\thanks{Cathy Wu is with the Massachusetts Institute of Technology, Cambridge, MA 02139, USA.
\tt\small{cathywu@mit.edu}}}

\maketitle

\begin{abstract}

Autonomous vehicles (AVs) are becoming increasingly popular, with their applications now extending beyond just a mode of transportation to serving as mobile actuators of a traffic flow to control flow dynamics. This contrasts with traditional fixed-location actuators, such as traffic signals, and is referred to as \textit{Lagrangian traffic control}. However, designing effective Lagrangian traffic control policies for AVs that generalize across traffic scenarios introduces a major challenge. Real-world traffic environments are highly diverse, and developing policies that perform robustly across such diverse traffic scenarios is challenging. It is further compounded by the joint complexity of the multi-agent nature of traffic systems, mixed motives among participants, and conflicting optimization objectives subject to strict physical and external constraints. To address these challenges, we introduce Multi-Residual Mixture of Expert Learning (MRMEL), a novel framework for Lagrangian traffic control that augments a given suboptimal nominal policy with a learned residual while explicitly accounting for the structure of the traffic scenario space. In particular, taking inspiration from residual reinforcement learning, MRMEL augments a suboptimal nominal AV control policy by learning a residual correction, but at the same time dynamically selects the most suitable nominal policy from a pool of nominal policies conditioned on the traffic scenarios and modeled as a mixture of experts. We validate MRMEL using a case study in cooperative eco-driving at signalized intersections in Atlanta, Dallas Fort Worth, and Salt Lake City, with real-world data-driven traffic scenarios. The results show that MRMEL consistently yields superior performance—achieving an additional 4\%–9\% reduction in aggregate vehicle emissions relative to the strongest baseline in each setting.

\end{abstract}


\section{INTRODUCTION}

Transportation systems are undergoing a technology-driven reconfiguration, fueled by the emergence of autonomous vehicle (AV) technologies. Beyond the role of AVs as mobility providers, they are now being studied as mobile actuators embedded within traffic flows, capable of shaping the collective behavior of surrounding vehicles through their local interactions~\cite{wu2021flow}. This paradigm, known as \textit{Lagrangian traffic control}, departs from traditional, fixed-location interventions (e.g., traffic signals) and has shown promising results in reducing congestion~\cite{wu2021flow}, improving safety~\cite{hickert2023cooperation}, and lowering emissions~\cite{meng2020eco} at the traffic flow level.

\begin{figure*}
    \centering
    \includegraphics[width=0.8\linewidth]{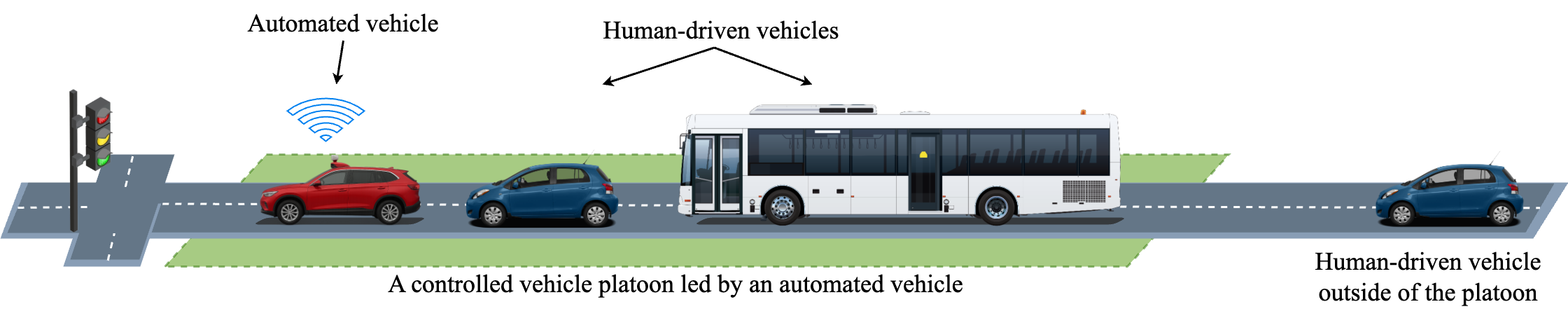}
    \caption{An illustration of Lagrangian traffic control at a signalized intersection, where a red AV optimizes its acceleration and lane changes to guide platoons of vehicles via car-following dynamics (e.g., to reduce fleet emissions). Multiple AVs can coordinate locally—e.g., driving side-by-side—to minimize disruptions from human drivers due to lane changes. }
    \label{fig:illustrative_intro}
\end{figure*}

At the heart of this approach lies a deceptively simple idea: AVs can modulate traffic by influencing the car-following behavior of nearby vehicles—both human and other AVs. Yet in practice, this gives rise to a richly complex control problem. The resulting system is a mixed traffic (i.e., both human-driven vehicles and AVs coexist), multi-agent environment characterized by mixed-motive interactions, non-stationary dynamics stemming from human behavioral and traffic scenario variations and multiple and conflicting objectives (e.g., efficiency vs safety). Despite this, much of the existing literature reduces the problem to narrowly scoped instances—optimizing isolated vehicles, minimizing single-objective cost functions, or assuming idealized traffic conditions such as homogeneous driver behavior or a few select traffic scenarios.

These simplifications, while analytically convenient and reasonable for a proof of concept, obscure the structural challenges inherent to real-world Lagrangian control. We observe that the overall complexity of this problem is largely stemming from the need for the control algorithms to generalization across diverse traffic scenarios which makes the other interlocking challenges, such as the complexity of multi-agent dynamics, the trade-offs between competing objectives, and the hard operational constraints, more challenging to handle. 

Conventionally, heuristics~\cite{katsaros2011performance}, model-based optimization methods such as dynamic programming, model predictive control~\cite{sajadi2019nonlinear}, and model-free learning-based methods such as reinforcement learning~\cite{wu2021flow} have been leveraged to develop Lagrangian control policies often under stylized assumptions that strip away the very complexities that make the problem realistically challenging. This raises a critical question: To what extent does the tractability of existing formulations come at the expense of real-world viability? We posit that this gap—between narrow theoretical contributions and the messy, multifaceted nature of real deployments—may be one of the key technical reasons why many Lagrangian traffic control strategies have yet to transcend the boundaries of academic research and achieve meaningful impact in the real world.

In this work, we take a step toward answering this question by introducing a novel learning framework: Multi-Residual Mixture of Experts Learning (MRMEL)—a framework designed to enable cooperative control of a fleet of AVs for Lagrangian traffic control. MRMEL unifies foundational ideas from residual reinforcement learning~\cite{silver2018residual, sergey_rpl} and multi-task learning within a mixture-of-experts architecture, purpose-built for generalizable multi-vehicle control. Central to the approach is a decomposition of the vehicle control policy into two synergistic components: a nominal policy, which provides a nominal control action derived from prior knowledge policy, and a residual function, which is trained to adaptively refine and correct this nominal in response to real-time observations. This residual formulation allows the learning process to focus on compensating for deficiencies in the nominal policy rather than solving the full control problem from scratch.

A key insight underpinning MRMEL is the observation that a given nominal policy behavior is not static across all traffic scenarios. Heterogeneity in traffic dynamics—stemming from variations such as flow patterns, road topologies, and the stochastic behavior of human drivers- demands a more careful treatment of prior knowledge in the form of a nominal policy. To that end, MRMEL incorporates a gated mixture of nominal policies: an ensemble of candidate prior policies, each potentially specialized for a different subset of traffic scenarios or states. A learned gating network dynamically modulates the influence of each expert based on a given traffic scenario. 

We empirically evaluate MRMEL in a large-scale, high-fidelity simulation study focused on cooperative eco-driving—a representative use case of Lagrangian traffic control aimed at reducing carbon emissions in mixed traffic settings. Our experiments are calibrated to reflect real-world traffic conditions across three major US metropolitan cities: Atlanta, Dallas-Fort Worth, and Salt Lake City. We benchmark MRMEL against several baselines, including multi-task policies trained from scratch, industry-standard eco-driving algorithms, and state-of-the-art residual reinforcement learning variants. Across all three cities and under two eco-driving adoption levels (30\% and 100\% of fleet vehicles are AVs), MRMEL consistently yields superior performance—achieving an additional 4\%–9\% reduction in aggregate vehicle emissions relative to the strongest baseline in each setting.

\section{Related Work}

\subsection{Lagrangian Traffic Control}

Lagrangian traffic control has recently attained significant attention in the robotics, machine learning, and intelligent transportation systems communities. The central idea is to utilize a subset of controlled AVs to influence overall traffic flow by indirectly affecting the behavior of surrounding human drivers through car-following dynamics. Wu et al.~\cite{wu2021flow} introduced Flow, a modular reinforcement learning framework for Lagrangian traffic control, which sparked subsequent research in this area. Building on Flow, several studies have explored diverse applications, including eco-driving~\cite{jayawardana2024generalizing}, congestion mitigation~\cite{poudel2024endurl}, and emergency vehicle path clearance~\cite{suo2024model}. In parallel, theoretical analyses have emerged, with a particular focus on the string stability of these mixed traffic systems~\cite{wu2018stabilizing}. More recently, a relatively large-scale field experiment with a hundred AVs has been conducted by Jang et al.~\cite{jang2025reinforcement} to demonstrate the real-world effectiveness of Lagrangian traffic control in smoothing highway traffic.

Despite this promising progress, most simulation-based studies remain confined to a limited set of traffic scenarios and often overlook the critical aspect of generalization—an essential requirement for real-world deployment. Existing real-world implementations typically focus on relatively static settings, such as specific highway segments, with control strategies tailored to these fixed environments~\cite{jang2025reinforcement}. In this work, we seek to bridge this gap by proposing a deep reinforcement learning framework capable of learning control policies that generalize across a wide range of traffic scenarios.

\subsection{Residual Reinforcement Learning and related concepts}

Residual Reinforcement Learning (RRL) was independently introduced by Johannink et al.\cite{sergey_rpl} and Silver et al.\cite{silver2018residual} for robotic control. RRL concerns with augmenting an arbitrary initial policy by learning a residual corrective function. Silver et al. emphasized RRL’s ability to improve policy learning under sensor noise, model inaccuracies, and controller miscalibrations, while Johannink et al. further highlighted its effectiveness for sim-to-real transfer. Since then, RRL has been applied to broader domains, including autonomous driving~\cite{shi2024task}, racing~\cite{zhang2022racing}, eco-driving~\cite{jayawardana2024generalizing}, and robotic manipulations~\cite{hao2022metarpl}.

Another related line of work explores combining model-based and model-free methods in distinct stages of training~\cite{hewing2020learning}. A common approach involves first learning a dynamics model, which is then used for downstream planning tasks~\cite{mordatch2016combining, kollar2018mpc}. Alternatively, some methods learn the dynamics model, value function, and policy jointly in an integrated framework~\cite{heess2015learning}.

Overall, while RRL has shown promise in single-agent, single-task settings—especially in robotic manipulation—its ability to generalize across diverse task variations remains underexplored. Similar limitations are seen in hybrid model-based and model-free approaches. This work aims to address this gap by proposing an extension to RRL to make it generalize better. 

\section{Preliminaries}

In the following subsections, we briefly introduce the related core concepts that are used in this work.

\subsection{Reinforcement Learning}
\label{rl}

Reinforcement learning enables learning an optimal control policy through interaction with an environment. This learning process is commonly modeled as a Markov Decision Process (MDP), defined by the tuple $M = \left\langle\mathcal{S}, \mathcal{A}, p, r, \rho, \gamma \right\rangle$. Here, $\mathcal{S}$ represents the set of possible states, $\mathcal{A}$ the set of available actions, and $p(s_{t+1} \mid s_t, a_t)$ denotes the probability of transitioning to state $s_{t+1}$ from state $s_t$ after taking action $a_t$. The reward function $r(s_t, a_t) \in \mathbb{R}$ quantifies the immediate feedback received after executing an action, while $\rho$ specifies the distribution over initial states. The discount factor $\gamma \in [0, 1]$ governs the trade-off between immediate and future rewards.

Given the MDP, the objective is to determine an optimal policy $\pi^*: \mathcal{S} \rightarrow \mathcal{A}$ over a finite horizon $H$ that maximizes the expected cumulative discounted reward. Formally, the optimal policy is defined as,
\vspace{-0.02cm}
\begin{equation}
\pi^* = \argmax_{\pi} \mathbb{E} \left[\sum_{t=0}^{H} \gamma^t r(s_t, a_t) \mid s_0 = s, \pi\right]
\end{equation}

\subsection{Residual Reinforcement Learning}
\label{residual-rl}

RRL is a framework that builds on standard reinforcement learning introduced in Section~\ref{rl} by incorporating prior knowledge in the form of an existing nominal policy into the learning process. Instead of learning a policy from scratch, RRL aims to learn a residual function that refines a predefined nominal policy. Formally, the agent’s overall action is a combination of the nominal policy $\pi_n$ and a learned residual function $f_{\theta}$ parameterized by $\theta$, typically expressed as, 

\begin{equation}
\label{rrl-equation}
\pi^*(s) = \pi_n(s) + f_{\theta}(s)
\end{equation}

Intuitively, when the nominal policy $\pi_n$ is close to optimal, the residual function only needs to make minor adjustments to the nominal control to achieve optimal performance. Even if $\pi_n$ is suboptimal, it can still serve as a useful prior, guiding exploration during the learning of the residual function.

\subsection{Multi-task Reinforcement Learning}
\label{multi-task-learning}

Multi-task reinforcement learning generalizes the single-task setting described in Section~\ref{rl} to a collection of tasks, each modeled as a distinct MDP. The goal is to learn a unified policy that performs well across all tasks in a set $\mathcal{T}$. Formally, the objective becomes,
\begin{equation}
\label{eq_mtl}
\pi^* = \argmax_{\pi} \mathbb{E} \left[\sum_{\tau \in \mathcal{T}} \sum_{t=0}^{H} \gamma^t r_{\tau}(s_t, a_t) \mid s_0=s, \pi \right]
\end{equation}

Here, $\mathcal{T}$ denotes the set of MDPs (tasks), and $r_{\tau}(s_t, a_t)$ is the reward function associated with task $\tau$. 

\subsection{Mixture of Experts}
\label{moe}

Mixture of Experts (MoE) is a modular learning paradigm that combines multiple specialized policies to solve complex tasks more effectively. Typically, each policy is trained or designed to handle specific regions of the input space, while a gating network dynamically assigns inputs to the most appropriate policy(s) based on relevance.

Formally, given a set of $K$ policies ${\pi_1, \pi_2, \dots, \pi_K}$ and a gating function $g: \mathcal{S} \rightarrow \Delta^K$ (a probability distribution over policies), the overall policy is defined as,
\begin{equation}
\pi(s) = \sum_{i=1}^K g_i(s) \pi_i(s)
\end{equation}

MoE promotes specialization, scalability, and interpretability, and is particularly effective in settings where diverse strategies may be required across different states or tasks.

\section{Method}

In this section, we motivate and formalize the requirement of generalization in Lagrangian traffic control and introduce our Multi-residual Mixture of Experts Learning (MRMEL) framework. In the following sections, we use cooperative eco-driving as the running example, though the framework is applicable to any Lagrangian traffic control task.

\subsection{Motivation}

The need for generalization in Lagrangian traffic control stems from the wide range of factors that influence vehicle behavior. In cooperative eco-driving, for instance, studies have identified approximately 33 factors that impact emission outcomes, leading to a high-dimensional problem space that suffers from the curse of dimensionality~\cite{jayawardana2024mitigating}. As a result, most existing approaches focus on a limited subset of these factors. This simplification is often motivated by the otherwise exponentially growing state space and the resultant complex expectation term in the optimization objective, which could lead to a challenging objective landscape. 

In MRMEL, our goal is to tackle the challenge of both policy and algorithmic generalization in Lagrangian traffic control. Policy generalization refers to a learned policy’s ability to perform effectively across a variety of environments, including those not encountered during training. This focus differs from algorithmic generalization, which concerns whether the learning algorithm itself can consistently discover effective policies in new and unseen settings. In our setting, MRMEL is designed to display policy generalization within each city and algorithmic generalization across cities (as they have different traffic scenario distributions). Moreover, we specifically address intra-task generalization, where the different environments arise from task variations within the same overarching task. In the context of cooperative eco-driving, such task variations stem from changes in factors such as lane length, road slope, or atmospheric temperature in each traffic scenario. Although the proposed method has the potential to be extended to inter-task generalization in building generally capable agents, where generalization spans across entirely different tasks, we leave that direction for future work.

\subsection{Problem Formulation}

We formalize a collection of task variations arising from a given task as a Contextual Markov Decision Process (CMDP)~\cite{hallak2015contextual}.
A CMDP is built on the MDP definition introduced in Section~\ref{rl} by using a \textit{context}—a set of parameters that characterize variations in the environment. Unlike the state of an MDP, which captures the step-by-step changes in the environment, context enables reasoning about the state evolution without needing direct access to the true transition and reward functions~\cite{benjamins2022contextualize}. Additionally, context features are generally static (i.e., they remain unchanged throughout an episode) or vary at a much slower rate than states. 

Mathematically, a CMDP can be defined as $\mathcal{M} = \left\langle\mathcal{S}, \mathcal{A}, \mathcal{C}, p_c, r_c, \rho_c, \gamma \right\rangle$. A notable difference from MDPs, is the introduction of a context space $\mathcal{C}$. The action space $\textit{A}$ and state space $\textit{S}$ remain unchanged. The transition dynamics $\textit{p}_c$, reward function $\textit{r}_c$, and initial state distribution $\rho_c$ are all conditioned on the context $c \in \mathcal{C}$, and vary across different task variations. Given a context $c$, the CMDP $\mathcal{M}$ is restricted to an MDP $\mathcal{M}_c$. It is then called a \textit{context-MDP} and the CMDP then manifests as a collection of context-MDPs\footnote{By MDP, we generally refer to any form of MDP, including but not limited to Partially Observable MDPs.}. 

More generally, we seek to find a set of at most $k$ policies $\Pi_k^{*}$ that maximizes the overall expected return across all context-MDPs within a CMDP where $\mathcal{R}(\pi, \mathcal{M}_c)$ is the expected return of policy $\pi$ on context-MDP $\mathcal{M}_c$,

\begin{equation}
\Pi_k^{*} = \argmax_{\Pi_k : |\Pi_k| \leq k} \mathbb{E}_{c \sim \rho(c)}\left[ \max_{\pi \in \Pi_k} \mathcal{R}(\pi, \mathcal{M}_c) \right]
\label{Eq:formalism}
\end{equation}

In this work, our goal is to achieve policy generalization across the entire context space of a given city and algorithmic generalization across cities. This leads to learning a single unified policy (\(k = 1\)) per city. While deploying multiple policies within a city may not stand as scalable, due to significant engineering overhead and increased system complexity, maintaining one policy per city remains a practical and justifiable compromise. It balances the need for generalization performance with the realities of deployment on resource-constrained onboard processors in AVs.

\subsection{Multi-residual Mixture of Experts Learning}
\label{mrmel}

\begin{figure*}
    \centering
    \includegraphics[width=0.9\linewidth]{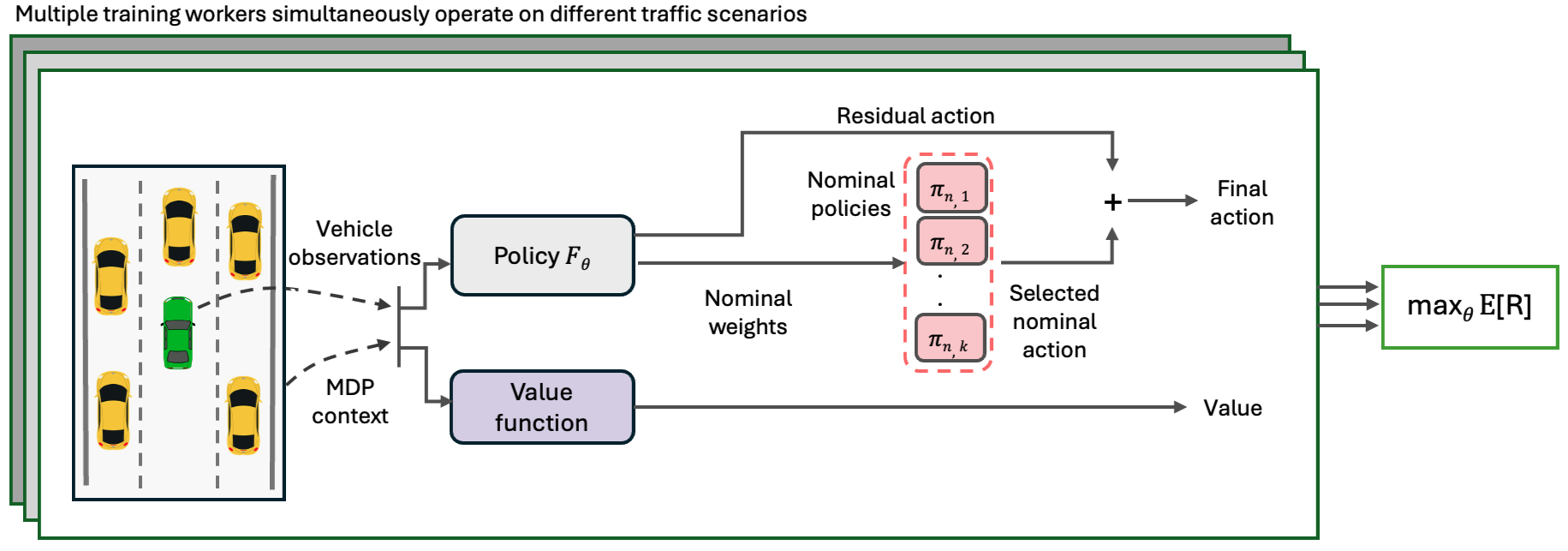}
    \caption{Schematic overview of the proposed method. Multiple training threads run in parallel, each sampling traffic scenarios from a predefined distribution. Within each thread, several AVs are simulated in a decentralized fashion, all sharing the same policy. Each AV independently observes its environment along with a context vector representing the current traffic scenario. These observations are fed into a policy $F_{\theta}$, which outputs a residual acceleration and a set of weights corresponding to a set of nominal policies. The final nominal acceleration is computed as a weighted sum of the nominal acceleration using these weights. The nominal acceleration is then added to the residual acceleration to produce the AV's final acceleration. }
    \label{fig:method-overview}
\end{figure*}

Learning a control policy from scratch that generalizes across the context space of a city is challenging due to many context variations and the general sensitivity of learning methods. To address this, we introduce \textit{Multi-residual Mixture of Expert Learning (MRMEL)}, a generic learning framework for learning policies that generalize. A visual illustration of the MRMEL framework is given in Figure~\ref{fig:method-overview}.

The key idea behind MRMEL is to extend the RRL as introduced in Section~\ref{residual-rl} to generalize better. Given that learning a policy from scratch can be particularly challenging when dealing with task variations, leveraging prior knowledge of the problem in the form of a nominal policy to warm-start the training process becomes a promising approach. To this end, we extend the RRL framework in two key directions.

To explicitly incorporate context variations into learning, we extend RRL with multi-task learning (Section~\ref{multi-task-learning}). Specifically, we leverage explicit context of a traffic scenario by conditioning the residual function not only on the state of the AV but also on a given context $c$, transforming the residual function into $f_\theta(s, c)$. However, the nominal policy does not necessarily need to be conditioned on the context, allowing practitioners the flexibility to determine the most suitable approach depending on the availability of nominal policies. This results in the RRL policy introduced in Equation~\ref{rrl-equation} taking a slightly different form, 

\begin{equation}
\pi^*(s, c) = \pi_n(s) + f_\theta(s, c)
\end{equation}

We note that the gradient of the $\pi^*$ does not depend on the $\pi_n$, enabling flexibility with nominal policy choice. The effectiveness of $\pi_n$ in a given context can vary depending on the nominal policy design and the system dynamics induced by the context. Hence, the role that the residual function $f_\theta$ has to play in each context variation varies. For instance, in cooperative eco-driving, the commonly used industry standard heuristic GLOSA policy performs effectively in traffic scenarios with low congestion~\cite{katsaros2011performance}. In these situations with GLOSA as the nominal, the residual function plays a minimal role. However, in congested traffic, the contribution of residual functions becomes significant as GLOSA underperforms. 

The above observation opens up an opportunity to improve the method further. In subspaces of the context space where one nominal performs poorly, we can leverage another nominal policy to improve the overall performance. However, it might be challenging to predetermine which parts of the context space could benefit from which nominal. Therefore, while addressing the limitation of using a single nominal policy for all task variations as introduced earlier, we further introduce a set of nominal policies using a mixture of expert architecture introduced in Section~\ref{moe}. Specifically, let there be \( K \) nominal policies \( \pi_{n,1}, \pi_{n,2}, \ldots, \pi_{n,K} \), where each nominal policy \( \pi_{n,k}(s) \) is specialized for a subspace of the context space. A gating mechanism \( g: (s, c) \to \Delta^{K-1} \) determines the weighting of each nominal policy for a given state \( s \) and context \( c \), and both soft and hard gating can be leveraged. The final policy is then defined as:
\[
\pi^*(s, c) = \sum_{k=1}^K g_{\rho}^k(s, c) \pi_{n}^k(s) + f_\theta(s, c)
\]
where \( g_{\rho}^k(s, c) \) represents the learned contribution of the \( k \)-th nominal policy \( \pi_{n}^k \) with the function $g$ parameterized by $\rho$.

The inclusion of a mixture of nominal policies allows flexibility in selecting the most useful nominal policy for each context based on the specific context characteristics. Further, note that one such nominal policy could be a zero-action policy (e.g., the policy that outputs zero acceleration in eco-driving), which allows the flexibility for the residual function to learn the full action instead of a residual action (since the effect of the nominal action is inactive in this setting). This feature is especially beneficial in contexts where all nominal policies are highly suboptimal (suboptimal than a random policy), and relying on them could make the residual learning process more challenging than learning from scratch.

\begin{table*}[h!]
    \centering
    \begin{tabular}{|c|cc|cc|cc|cc|cc|cc|}
        \hline
        \multirow{2}{*}{Method} & \multicolumn{4}{c|}{Dallas} & \multicolumn{4}{c|}{Atlanta} & \multicolumn{4}{c|}{Salt Lake City} \\
        \cline{2-13}
        & \multicolumn{2}{c|}{30\%} & \multicolumn{2}{c|}{100\%} & \multicolumn{2}{c|}{30\%} & \multicolumn{2}{c|}{100\%} & \multicolumn{2}{c|}{30\%} & \multicolumn{2}{c|}{100\%} \\
        & Emis. $\uparrow$ & Thr. $\uparrow$ & Emis. $\uparrow$ & Thr. $\uparrow$ & Emis. $\uparrow$ & Thr. $\uparrow$ & Emis. $\uparrow$ & Thr. $\uparrow$ & Emis. $\uparrow$ & Thr. $\uparrow$ & Emis. $\uparrow$ & Thr. $\uparrow$ \\
        \hline
        GLOSA & 0.75\% & 0.92\% & 1.31\% & 3.58\% & 1.58\% & 1.02\% & 3.71\% & 3.81\% & 1.02\% & 1.80\% & 1.49\% & 3.44\% \\
        Multi-task learning & 0.05\% & 0.0\% & 0.00\% & 0.00\% & 0.11\% & 0.0\% & 0.00\% & 0.00\% & 0.00\% & 0.00\% & 2.33\% & 2.71\% \\
        MRTL & 1.93\% & 2.50\% & 1.26\% & 4.23\% & 1.36\% & 1.23\% & 2.58\% & 3.90\% & 1.39\% & 2.32\% & 1.50\% & 4.37\% \\
        RRL (IDM) & 1.40\% & 1.44\% & 0.58\% & 0.51\% & 2.50\% & 1.71\% & 3.90\% & 3.90\% & 1.72\% & 2.01\% & 1.16\% & 0.60\% \\
        RRL (Const. acc.) & 0.05\% & 0.00\% & 0.00\% & 0.01\% & 0.05\% & 0.00\% & 8.66\% & 4.19\% & 0.05\% & 0.00\% & 0.00\% & 0.01\% \\
        RRL (Const. dec.) & 0.05\% & 0.00\% & 0.00\% & 0.01\% & 0.11\% & 0.00\% & 0.00\% & 0.01\% & 0.05\% & 0.00\% & 0.00\% & 0.01\% \\
        \textbf{MRMEL (Ours)} & \textbf{5.90\%} & \textbf{3.89\%} & \textbf{9.98\%} & \textbf{11.81\%} & \textbf{6.43\%} & \textbf{2.68\%} & \textbf{14.40\%} & \textbf{8.80\%} & \textbf{6.31\%} & \textbf{4.70\%} & \textbf{10.32\%} & \textbf{7.73\%} \\
        \hline
    \end{tabular}
    \caption{Performance of eco-driving methods across Dallas, Atlanta, and Salt Lake City at 30\% and 100\% AV penetration.}
    \label{tab:combined-results}
\end{table*}

\section{MRTL for Cooperative Eco-driving}

In the following subsections, we introduce cooperative eco-driving at signalized intersections as a case study for evaluating MRMEL for Lagrangian traffic control at the scale of metropolitan cities. 

\subsection{Cooperative Eco-driving CMDP}

Cooperative eco-driving aims to reduce vehicle exhaust emissions in mixed traffic environments while minimizing disruptions to overall travel time. The central idea is to control a set of AVs in a way that subtly influences human drivers—encouraging more environmentally friendly driving behaviors through car-following dynamics. AVs can collaborate by forming local teams that guide surrounding human drivers toward more efficient driving patterns. We focus specifically on cooperative eco-driving at signalized intersections, where numerous context factors—such as traffic signal timing, lane configurations, speed limits, and vehicle arrival rates—significantly influence emissions and, consequently, the effectiveness and generalizability of control policies. These factors collectively define the context for eco-driving CMDPs.

More formally, given an instantaneous emission model $E(\cdot)$ that measures vehicular emission, we seek an AV control policy such that, 

\vspace*{-0.05cm}
\begin{equation}
\label{eco-drive-objective}
\pi^* = \argmin_{\pi} \mathop{\mathbb{E}} \left[\sum_{c \in \mathcal{C}} \sum_{i=1}^{n_c} \int_{0}^{T_i} E\left(a_i(t), v_i(t)\right) dt + T_i \right]
\end{equation}

Here, $n_c$ represents the total number vehicles in each traffic scenario defined by context $c$, $v_i(t)$, and $a_i(t)$ denote the instantaneous speed and acceleration of vehicle $i$ at time $t$, $T_i$ denotes the travel time of vehicle $i$ and $\mathcal{C}$ is the set of traffic scenarios.

We define each context-MDP with context $c$ as follows, 

\begin{itemize}
\item \textbf{States}: Each vehicle observes its own position and speed, as well as those of the leading and following vehicles in the same and adjacent lanes. It also receives information about the current traffic signal phase and the remaining time until the next change. Additionally, context features include the full traffic signal plan (green and red phase durations), speed limit, lane length, road grade, vehicle type, engine type, and vehicle age.
\item \textbf{Actions}: longitudinal accelerations of the vehicle. 
\item \textbf{Rewards}: The reward at time step $t$ is defined as $r(t) = v(t) - w_1 e(t) - w_2 s(t) - w_3 a(t)$, where $v(t)$ is the velocity, $e(t)$ is the emission, $s(t)$ is a penalty for vehicle stops, and $a(t)$ is the absolute acceleration which are weighted to reflect multiple objectives: minimizing travel time, reducing emissions, discouraging idling, and smooth acceleration for passenger comfort. The hyperparameters are set to $w_1=30, w_2=15$, and $w_3=10$.

To encourage both cooperative and individual behavior, we adopt a stochastic reward assignment mechanism. At every step, with probability $p$, the reward is computed based on fleet-level metrics and shared uniformly among all AVs. With probability $1 - p$, each vehicle receives a reward based on its own behavior. This approach promotes cooperation while preserving individual autonomy when needed. In our experiments, we set $p = 0.2$.
\end{itemize}

In this work, we utilize pre-built eco-driving CMDPs from IntersectionZoo~\cite{jayawardana2024intersectionzoo}, a dataset of real-world, intersection-based eco-driving scenarios calibrated using 33 key influencing context factors. For further details on these scenarios and underlying context distributions, we refer the reader to the Jayawardana et al.~\cite{jayawardana2024intersectionzoo}.

\begin{figure*}[h]
    \centering
    \includegraphics[width=0.9\linewidth]{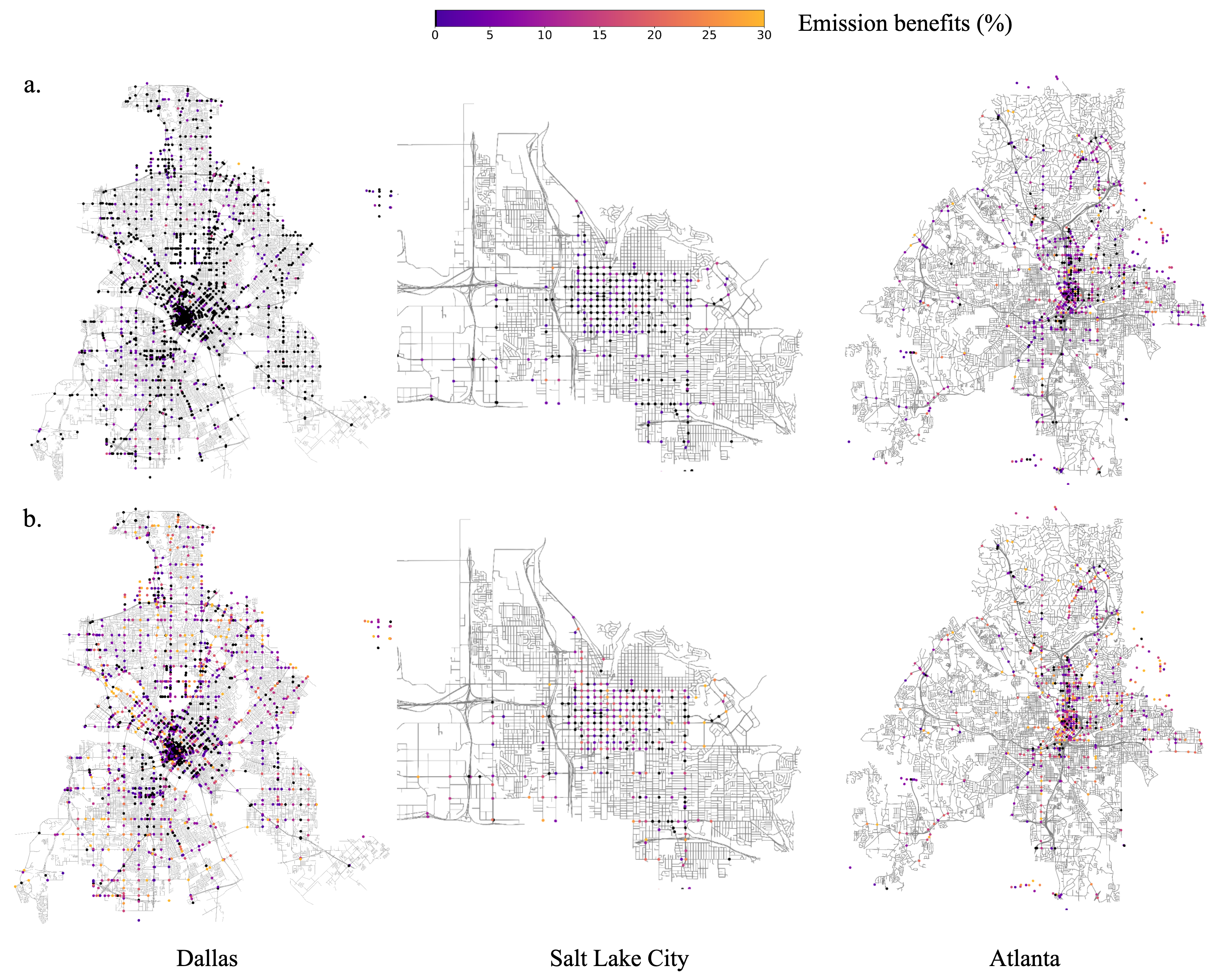}
    \caption{Spatial emission benefits distribution of all intersections in each city under 100\% AV penetration, color-coded by the emission benefit percentage. \textbf{a.} benefits from the best baseline method of each city and \textbf{b.} the benefits from the MRMEL. }
    \label{fig:spatial-benefits}
\end{figure*}

\subsection{Nominal Policies}
\label{nominal-policies}

We leverage five nominal policies in the MRMEL design for cooperative eco-driving as listed below. 

\begin{itemize}
    \item \textbf{GLOSA Controller}~\cite{katsaros2011performance}: A widely used eco-driving policy in the automotive industry that suggests accelerations based on a simple vehicle dynamics model and traffic signal timing. It does not account for surrounding vehicles and aims to minimize vehicle stops by gliding toward intersections when the signal is red. However, GLOSA tends to be suboptimal in dense traffic conditions where interactions with nearby vehicles are significant.
    
    \item \textbf{Constant Acceleration}: A basic heuristic policy that continuously applies a fixed acceleration of 0.1~$m/s^2$. 
    
    \item \textbf{Constant Deceleration}: A counterpart to the constant acceleration policy that applies a constant deceleration of 0.1~$m/s^2$. It reflects common eco-driving strategies that prioritize smooth deceleration to reduce energy consumption and avoid aggressive acceleration.
    
    \item \textbf{IDM (Intelligent Driver Model)~\cite{treiber2013traffic}}: A well-established car-following model that emulates naturalistic human driving behavior. It adjusts the ego vehicle's acceleration based on its current speed and the gap to the vehicle ahead, making it suitable for scenarios requiring human-like driving behaviors.
    
    \item \textbf{Zero-Action Policy}: A degenerate policy that always outputs zero acceleration (0~$m/s^2$). This serves as a baseline for the residual function, enabling it to learn complete acceleration from scratch in cases where all nominal policies are deemed insufficient.
\end{itemize}

\section{Experimental Results}

\subsection{Baselines}

We define three baselines and four variants of MRMEL to compare the performance of MRMEL. 

\begin{enumerate}
    \item \textbf{Intelligent Driver Model (IDM):} Used to compare emission benefits against human-like driving. We use the IDM~\cite{treiber2013traffic} model here due to its widespread adoption. 
    \item \textbf{Multi-task reinforcement learning:} a multi-task reinforcement learning-based baseline that learns a policy from scratch conditioned on the traffic scenario context.
    \item \textbf{GLOSA:} The industry standard GLOSA eco-driving controller~\cite{katsaros2011performance}.
\end{enumerate}

Additionally, we evaluate the effectiveness of the MoE architecture in MRMEL by comparing it against a set of it's own variants as baselines where each of the five nominal policies introduced in Section~\ref{nominal-policies} is used individually with residual reinforcement learning—i.e., without the MoE setup, relying instead on a single-policy residual learning approach. The variant that uses GLOSA as the nominal policy has been previously proposed by Jayawardana et al.~\cite{jayawardana2024generalizing} and is referred to as Multi-residual Task Learning (MRTL).

\subsection{MRMEL Implementation Details}

We adopt actor-critic architecture, as illustrated in Figure~\ref{fig:method-overview}, using a multi-layer perceptron with four hidden layers of 256 neurons each for both the actor and the critic. The actor (policy network) outputs both a residual action—sampled from a Gaussian distribution—and a one-hot encoding for nominal policy selection. We find that allowing the actor to jointly control both residual actions and nominal policy selection performs better than using a separate gating network for the latter. Training is conducted using the PPO algorithm~\cite{schulman2017proximal} with a learning rate of 0.0001.

Using the IntersectionZoo~\cite{jayawardana2024intersectionzoo} eco-driving dataset, we evaluate MRMEL across three major U.S. metropolitan cities. We simulate traffic scenarios under two levels of AV penetrations: 30\% and 100\%. In total, we model nearly 5,000 traffic scenarios across 1,670 intersections in Dallas–Fort Worth, 621 in Atlanta, and 282 in Salt Lake City.

\subsection{MRMEL Evaluations}

In Table~\ref{tab:combined-results}, we present the percentage improvements in emissions and intersection throughput achieved by MRMEL and other baseline methods, relative to human-like driving behavior modeled by IDM, under 30\% and 100\% AV penetration scenarios. Across all scenarios, MRMEL consistently and significantly outperforms the baselines in both metrics. Notably, higher AV penetration levels yield greater benefits, as a larger proportion of vehicles can adopt eco-driving strategies. 

We also observe that in some cases—such as Dallas at 100\% penetration—the RRL policy with a single nominal policy performs worse than GLOSA. This highlights the limitations of relying on a single policy and highlights the effectiveness of MRMEL’s mixture-of-experts framework, which adaptively fuses multiple nominal policies to optimize performance across diverse traffic scenarios. Moreover, we observe that MRTL~\cite{jayawardana2024generalizing} performs on par with, or occasionally worse than, GLOSA in certain cases. This highlights a key limitation in MRTL in real-world traffic scenarios, as MRTL has previously only been evaluated using synthetic traffic scenarios, which may not capture the real-world challenges appropriately.

Figure~\ref{fig:spatial-benefits} shows a spatial distribution of intersections in each city, color-coded by emission benefit percentages under 100\% AV penetration. In Figure~\ref{fig:spatial-benefits}a, we observe that the second-best baseline methods result in relatively low benefits across most intersections, with any improvements appearing scattered—potentially due to overfitting to specific intersection types that are easier to solve than others. In contrast, Figure~\ref{fig:spatial-benefits}b demonstrates that MRMEL not only achieves higher emission benefits overall but also extends these benefits more broadly across intersections. This suggests that MRMEL improves performance even at intersections that previously showed little to no benefit, highlighting its generalization capability.

\begin{figure}
    \centering
    \includegraphics[width=0.9\linewidth]{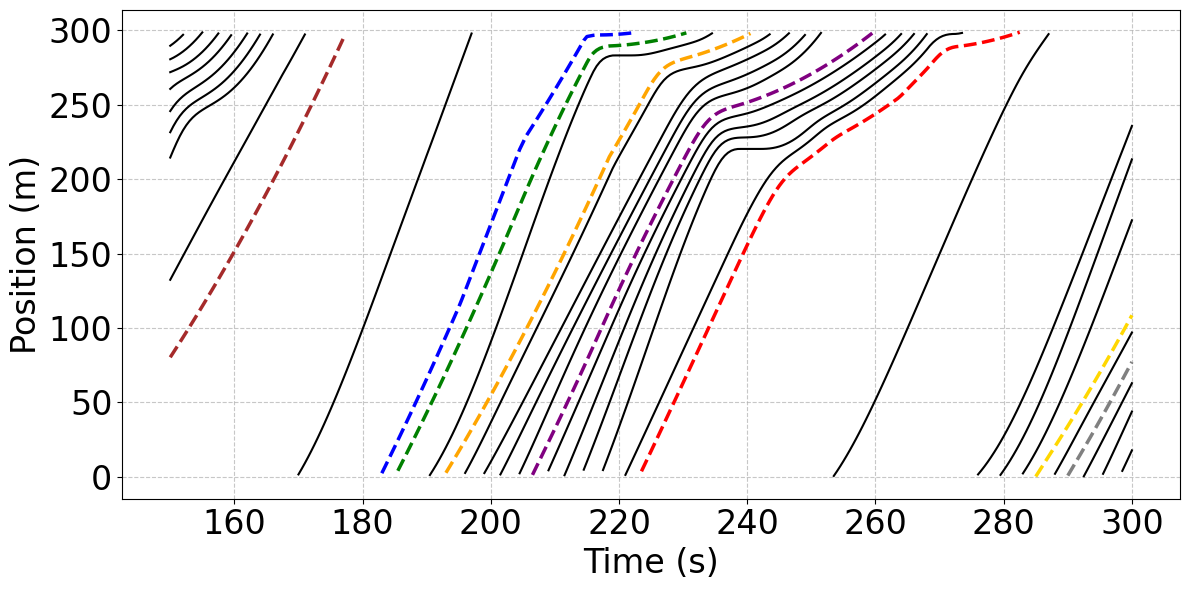}
    \caption{Time space diagram of a selected traffic scenario when the vehicles approach a signalized intersection (traffic signal at 300m mark). The black solid line indicates human-driven vehicles, and colored dashed lines indicate AVs. }
    \label{fig:time_space}
\end{figure}

In Figure~\ref{fig:time_space}, we present a time–space diagram for a selected traffic scenario to illustrate how MRMEL-driven vehicle behaviors help reduce emissions. Notably, we observe most AVs are gliding toward the intersection when the traffic signal is already red. This approach minimizes idling, which produces emissions without contributing to vehicle movement. Additionally, by gliding, AVs help form human-driven vehicle platoons, which can be then more effectively controlled to further reduce emissions at the fleet level.

\begin{figure}
    \centering
    \includegraphics[width=1.0\linewidth]{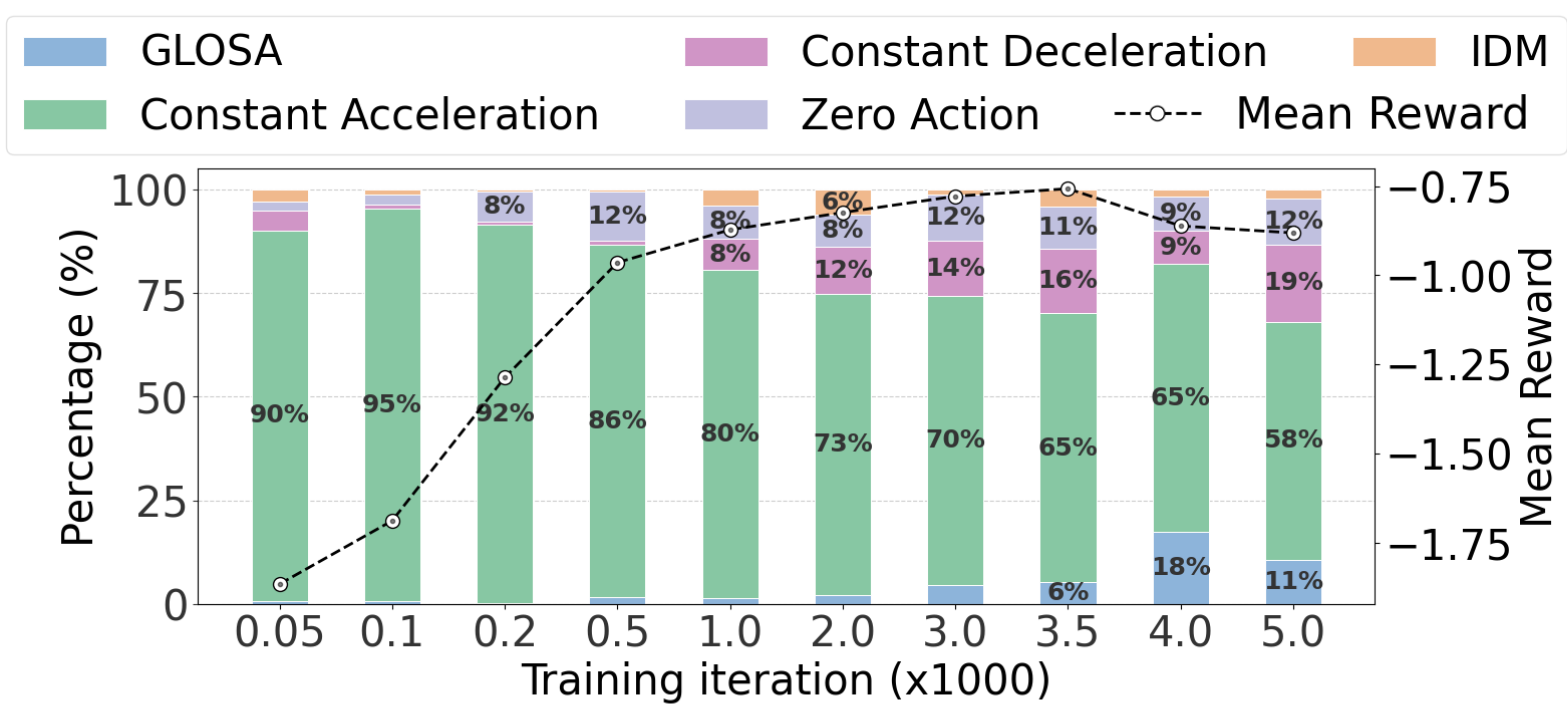}
    \caption{The percentage use of each nominal policy within the MRMEL framework during 5000 iterations of training. The mean reward of the training is given in dashed black line. }
    \label{fig:nominal_policy_use}
\end{figure}

Finally, Figure~\ref{fig:nominal_policy_use} shows the percentage usage of each nominal policy over 5,000 training iterations in training MRMEL. In the early stages of training, MRMEL predominantly relies on the constant acceleration nominal policy and changes to other policies in the latter stages of the training. This behavior may reflect an implicitly emerging curriculum during training. Intuitively, before learning to optimize for emissions reduction, the agent must first learn basic driving behavior—specifically, how to move forward, which is effectively supported by the constant acceleration policy.

As training progresses, the use of this policy declines by approximately 30\% from its peak at iteration 100 to iteration 3,500, which corresponds to the highest average reward. This suggests that once the model has learned how to drive, the second phase of the implicit curriculum begins: refining behavior to reduce emissions. Here, MRMEL increasingly utilizes the constant deceleration policy, likely to enable gliding, and the zero-action policy, which overrides the nominal behavior.

In summary, Figure~\ref{fig:nominal_policy_use} demonstrates how MRMEL leverages a mixture of nominal policies to potentially form an implicit curriculum. This curriculum guides the agent through progressively more complex behaviors, enabling more structured and effective exploration in reinforcement learning, leading to the development of a higher-performing overall policy.

\section{Conclusion}

In this work, we introduce Multi-Residual Mixture of Experts Learning (MRMEL)—a versatile framework that extends residual reinforcement learning by incorporating multi-task learning and a mixture of multiple nominal policies within a mixture of expert architecture, specifically applied to Lagrangian traffic control. Our empirical results demonstrate that MRMEL significantly improves the learning of cooperative eco-driving policies, highlighting its effectiveness. Currently, similar to standard residual reinforcement learning, MRMEL is limited to continuous control tasks. Extending the framework to handle discrete control remains an important direction for future work. Furthermore, due to its generality, MRMEL holds promise for broader applications in domains such as robotics, where strong generalization capabilities are particularly useful. We leave these extensions for future work.

\bibliographystyle{unsrt}
\bibliography{references}

\end{document}